\documentclass[11pt]{article}
\usepackage{paclic34}
\usepackage{times}
\usepackage{latexsym}
\usepackage{amsmath}
\usepackage{multirow}
\usepackage{url}

\usepackage[natbibapa]{apacite}
\usepackage{graphicx}
\usepackage{soul}
\usepackage{color}
\usepackage[table]{xcolor}% http://ctan.org/pkg/xcolor
\usepackage{pbox}
\usepackage{alltt}
\usepackage{framed}

\title{Predicting gender and age categories in English conversations using lexical, non-lexical, and turn-taking features}

\author{Andreas Liesenfeld, Gábor Parti, Yu-Yin Hsu, Chu-Ren Huang \\
  Department of Chinese and Bilingual Studies\\
  The Hong Kong Polytechnic University  \\
  Hong Kong \\
  {\tt amliese;gabor.parti;yu-yin.hsu;churen.huang@connect.polyu.hk}
  }

\date{}

\begin{document}
\maketitle
\begin{abstract}
  This paper examines gender and age salience and (stereo)typicality in British English talk with the aim to predict gender and age categories based on lexical, phrasal and turn-taking features. We examine the SpokenBNC, a corpus of around 11.4 million words of British English conversations and identify behavioural differences between speakers that are labelled for gender and age categories. We explore differences in language use and turn-taking dynamics and identify a range of characteristics that set the categories apart. We find that female speakers tend to produce more and slightly longer turns, while turns by male speakers feature a higher type-token ratio and a distinct range of minimal particles such as ``eh", ``uh'' and ``em". Across age groups, we observe, for instance, that swear words and laughter characterize young speakers' talk, while old speakers tend to produce more truncated words. We then use the observed characteristics to predict gender and age labels of speakers per conversation and per turn as a classification task, showing that non-lexical utterances such as minimal particles that are usually left out of dialog data can contribute to setting the categories apart. 
\end{abstract}

\noindent
\textbf{Author's note (Oct 2020): statement on the use of social categories in this study} 
\noindent
\textit{This work involves the labelling of participants for social categories related to gender and age. We caution against the use of this heuristic due to the risk of promoting a biased view on the topics. We would like to encourage those interested in the computational modelling of social categories to join the discussion on these concerns and consider participating in efforts to build more inclusive resources for the study of the topics.}

\section{Introduction}

One of the most interesting topics in language studies has been on how speakers' gender and age differences influence their communicative behaviour. Transcriptions of real-world, naturally-occurring conversations provide us a window to examine such differences in talk-in-interaction. 

Gendered and age-salient elements of talk have long been studied from a range of perspectives, including linguistics (e.g. \citealt{lakoff1973language}, \citealt{tannen1990you}), psychology (for an overview see, e.g. \citealt{james1993understanding}, \citealt{tannen1993gender}), and conversation analysis (e.g. \citealt{jefferson1988sequential}). This topic has also been extensively studied from a computational perspective, focusing on how the differences can be formally described and modelled. In recent years, the research interest has been extended to various applications using different types of data, such as using movie subtitles to identify gender distinguishing features \citep{schofield2016gender}; email interactions to study gender and power dynamics \citep{prabhakaran2017dialog}; video recordings of human-robot interactions to study gendered and age-related differences in turn-taking dynamics \citep{skantze2017predicting}; literary and weblog data to study differences between male and female writing (\citealt{herring2006gender}; \citealt{argamon2003gender}); and multimodal audiovisual and thermal sensor data for gender detection \citep{abouelenien2017multimodal}. 

Recent studies on gendered and age-salient behaviour in conversations also focus on the use of specific constructions or classes of constructions, such as swear words \citep{mcenery2004swearing}, amplifiers \citep{xiao2007corpus}, \textit{do} constructions \citep{oger2019study}, and minimal particles \citep{acton2011gender}. In the current study, we use transcriptions of recordings of naturally-occurring talk in British English to explore distributional differences in language use across gender and age groups, testing well-known tropes such as tendencies that women speak more politely, or that men use more swear words (\citealt{baker2014using}, \citealt{lakoff1973language}, \citealt{tannen1990you}), and also shedding light on other under-explored aspects of gendered and age-salient elements in talk such as the use of non-lexical vocalizations, laughter and other turn-taking dynamics. Our interest in this topic derives from work in computational modelling of dialog and conversation, especially studies aiming to automatically identify speaker properties for the use in voice technology and user modeling (\citealt{joshi2017personalization}, \citealt{wolters2009age}, \citealt{Liesenfeld}).

This pilot study explores whether non-lexical vocalizations and turn-taking properties can contribute to the prediction of age and gender categories. We investigate this question using a dataset of naturalistic talk that includes a range of elements other than words, such as laughter, pauses, overlaps and minimal particles. Can authentic and often ``disfluent" and ``messy" transcriptions of natural talk be used for a classification task? How will different behavioural cues contribute to a statistical investigation and prediction of gender and age salience and typicality?

\section{Data description}

Our dataset comes from the Spoken BNC2014 \citep{love2017spoken}, a corpus of contemporary British English conversations recorded between 2012-2016. It consists of transcriptions of talk on a range of topics covering everyday life in casual settings between around 2 to 4 speakers with a wide variety of social relationships such as between family members or good friends, and among colleagues or acquaintances. For classification, we extract two subcorpora from the SpokenBNC using the speaker labels ``female" and ``male" as well as age labels for speakers above 70 years and under 18 years. Table \ref{table:1} provides an overview of the two subcorpora. For age, we chose to only include the youngest and oldest speakers to tease out more significant differences by removing the bulk of middle-aged speakers. The downside of this approach is that this subcorpus is relatively small. 

\begin{table}[h]
\begin{center}
\begin{tabular}{|l|l|l|}
\hline \bf Feature & \bf Category & \bf Count \\ \hline
Speakers & Female & 365 \\ 
  & Male & 305 \\ 
  & Old  & 56 \\ 
   & Young  & 49 \\ \hline
Words  & Female & 6,671,774  \\ 
  & Male  & 4,080,524 \\ 
  & Old  & 737,398 \\ 
  & Young  & 792,039 \\
\hline
Turns  & Female & 742,973  \\ 
  & Male  & 478,851 \\ 
  & Old  & 96,994 \\ 
  & Young  & 102,433 \\
\hline
Average turn length & Female & 9.42  \\ 
(in words)  & Male  & 8.950 \\ 
  & Old  & 8.05 \\ 
  & Young  & 8.18 \\
\hline
Type-token ratio & Female & 0.0073  \\ 
  & Male  & 0.011 \\ 
  & Old  & 0.0231 \\ 
  & Young  & 0.0235 \\
\hline

\end{tabular}
\end{center}
\caption{Properties of the dataset obtained from the SpokenBNC2014 corpus. "Old" refers to speakers above 69 years of age, "young" includes speakers up to 18 years old.}
\label{table:1}
\end{table}

% An essential part of age and gender identification in such systems is to have a good understanding of features that characterize gendered and age-salient talk in a given speech community. To this end, we examine our data by two broad sets of categories of speakers: ``gender" (female and male) and ``age" (old and young). \hl{This arbitrary division of speakers by age -- labelled ``old" vs. ``young" -- , instead of using multiple categories based on age-range is an inconvenient necessity forced by the disproportion of the corpus, in favour of young adults (the age group 19-29 is represented by at least 2.5x more words than other age groups). Since we did not want to loose any valuable data... (((relative frequency=empirical probability)))} 

\section{Methods}

% \begin{table*}[!ht]
% \centering
% \begin{tabular}{|l|c|r|r|r|}
% \hline
% \multicolumn{1}{|c|}{\multirow{3}{*}{\textbf{Features}}} & \multicolumn{4}{c|}{\textbf{Category}} \\ \cline{2-5} 
% \multicolumn{1}{|c|}{} & \multicolumn{2}{c|}{\textbf{Female}} & \multicolumn{2}{c|}{\textbf{Male}} \\ \cline{2-5} 
% \multicolumn{1}{|c|}{} & \textbf{Old} & \multicolumn{1}{c|}{\textbf{Young}} & \multicolumn{1}{c|}{\textbf{Old}} & \multicolumn{1}{c|}{\textbf{Young}} \\ \hline
% Token counts & \multicolumn{1}{r|}{2598317} & 2380088 & 1780529 & 1222050 \\ \hline
% Type counts & \multicolumn{1}{r|}{38516} & 35853 & 36141 & 30509 \\ \hline
% Type-token ratio (TTR) & \multicolumn{1}{r|}{0.0148} & 0.0151 & 0.0203 & 0.0250 \\ \hline
% \end{tabular}
% \caption{\label{font-table} Types of lexical features and their results across gender and age categories}
% \label{table:2}
% \end{table*}

Comparing the behaviour of speakers across categories, we first look at lexical and phrasal differences. Then we examine non-lexical vocalizations such as laughter, minimal particles, and turn-taking dynamics such as overlaps and pauses. For both parts, we tokenize the corpora and remove stopwords using the NLTK and SpaCy libraries (\citealt{bird2009natural}, \citealt{honnibal2017spacy}).

\subsection{Lexical features}

We select a number $k_i$ of speaker's of each label from conversations $i$ and build a language model with the n-gram frequencies for all turns per category. We then examine the characteristic differences in the use of lexical items using Scaled F-score. Scaled F-score is a modified metric based on the F-score, the harmonic mean of precision and recall. It addresses issues related to harmonic means dominated by precision, as well as a better representation of low-frequency terms.\footnote{https://github.com/JasonKessler/scattertext} \\
We plot gender and age categories using the Scattertext library (\citealt{kessler2017scattertext}) to visualize the cross-categorical differences at the n-gram level. Figure 1 and 2 show words and phrases that are more characteristic of each category, while also reporting their frequencies and a list of top characteristic items.

\begin{figure*}[!ht]
\begin{center}
\includegraphics[width=\pdfpagewidth, angle=90]{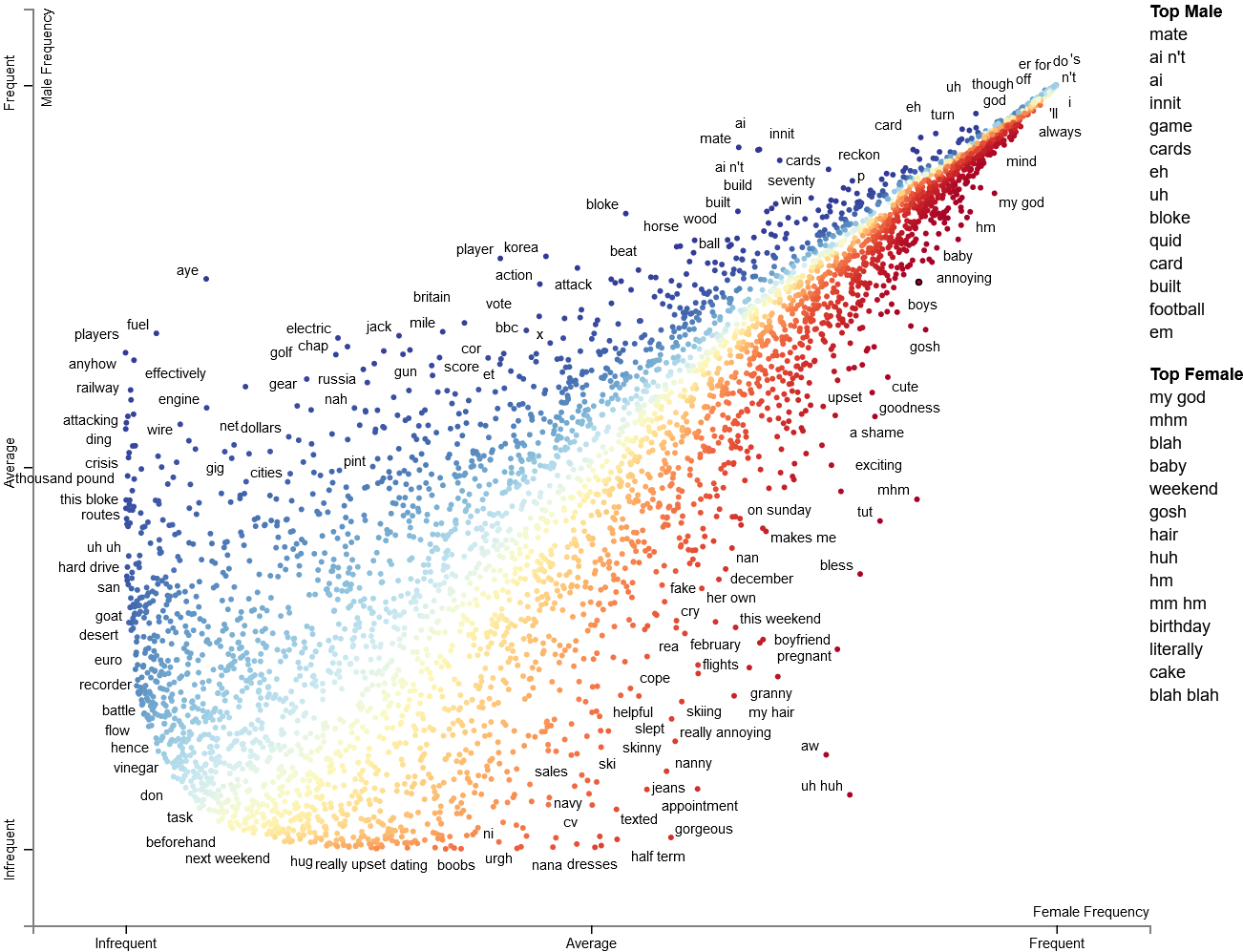}
\caption{Overview of characteristic terms by gender category, blue=Male, Red=Female, plotted by frequency}
\end{center}
\end{figure*}

\begin{figure*}[!ht]
\begin{center}
\includegraphics[width=\pdfpagewidth, angle=90]{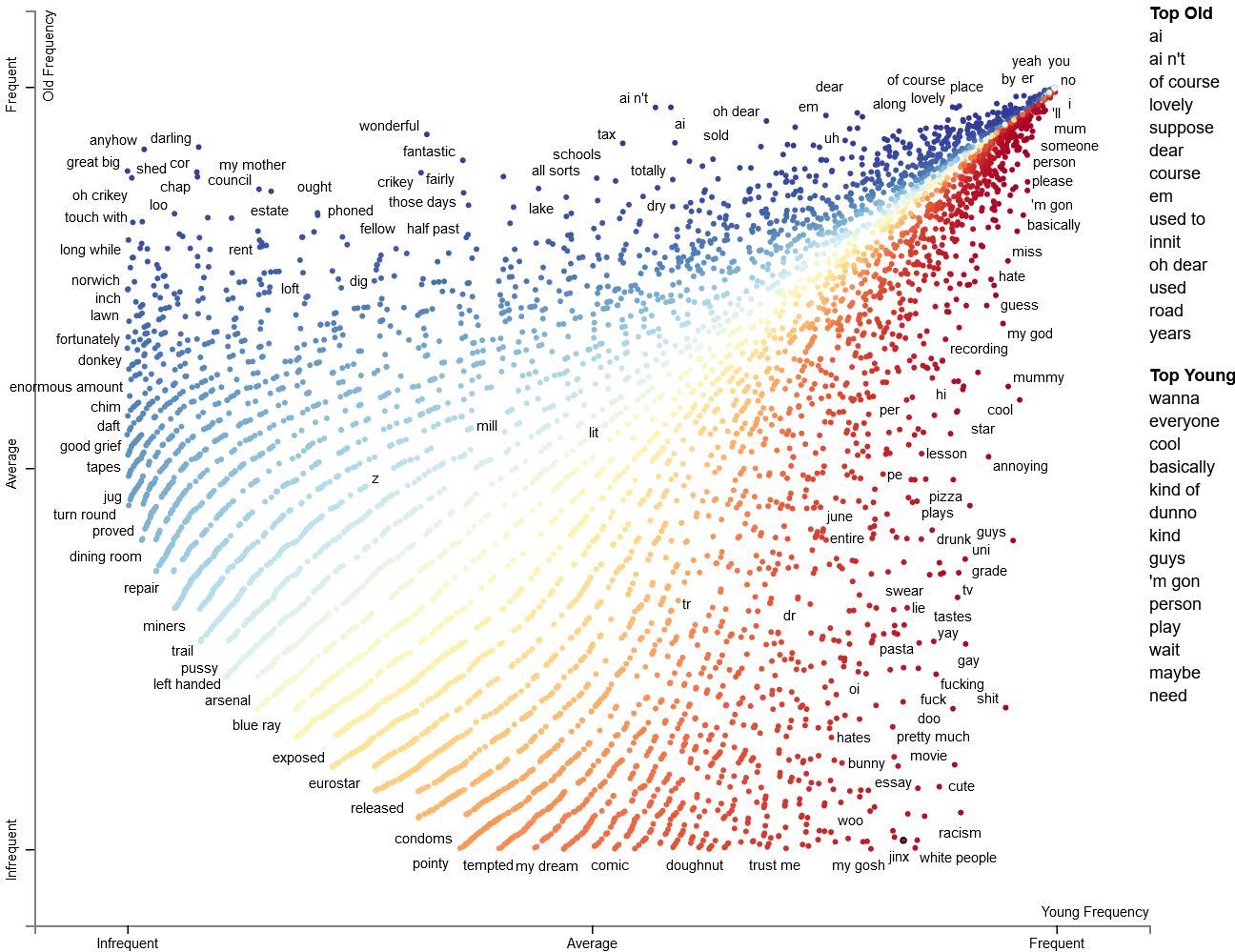}
\caption{Overview of characteristic terms by age category, blue=Old, Red=Young, plotted by frequency}
\label{fig.:1}
\end{center}
\end{figure*}

% \begin{figure}
% \includegraphics[width=0.5\textwidth]{unigram.png}
% \caption{Comparison clouds based on Top 600 unigram differences across gender and age groups (font-size mapped the maximum frequency deviation, and the centeredness ranked by probability)}
% \label{fig.:2}
% \end{figure}

% We also plot unigram differences of the four subcorpora as a wordcloud with font size increasing with the frequency deviation, and the tokens centralizing with increasing frequency (Figure \ref{fig.:2}). This allows a more broad-stroked comparison of the top tokens across the four categories. 
We observe that a range of terms reflect (stereo)typicality of gender and age categories in our corpus. For instance, top characteristic terms of male speakers feature the nouns ``mate", ``game", ``cards", ``quid" and ``football", while female speaker's talk more prominently features ``baby", ``weekend", ``hair", ``birthday" and ``cake" (see Figure 1). More interesting for us, the characteristic terms per gender category also feature a number of verb constructions, exclamations and minimal particles such as ``ain't", ``innit", ``eh" and ``uh" for male speakers and ``my God", ``mhm", ``blah". ``huh" and ``hm" for female.  

For age categories, notably a much smaller corpus, we also observe that a range of items features more prominently in talk of speakers labelled as young or old (see Figure 2). Likewise, we observe that some non-lexical utterances, exclamations and particles exhibit category salience, such as ``em", ``innit" and ``oh dear". Based on these observations we decide to further explore the role of non-lexical vocalizations, exclamations and minimal particles in the corpus.

\begin{framed}
\label{table2}
\begin{alltt}\small
\textbf{1 Positive responses and continuers:}
S1: you're so good at hair
S2: really?
S1: \textbf{mm}
S2: hair is my weakness I feel like 
    I'm really bad

S1: no I'd rather sit inside
S2: \textbf{uhu}
S1: if it was just a little bit 
    sunnier

\textbf{2 Turn stalling:} 
S1: you don't like riding them?
S2: I do but [short pause] \textbf{hmm}
    [short pause] you don't really
S3: [overlapping] I don't have a 
    bike

\textbf{3 Turn management:}
S1: oh lemon balm yeah you can do 
    that as well
S2: \textbf{erm} what what is very good for
    colds i [truncated] is \textbf{er erm}
    purple sage
S1: yeah pur [truncated] yeah
    [short pause] I know that one 
    yeah

\textbf{4 Repair initiators:}
S1: are all all the actors are 
    redubbed for the songs aren't 
    they
S2: \textbf{hm?}
S1: are the all the actors redubbed 
    for the songs? I can't remember

\textbf{5 Change-of-state tokens:}
S1: she was just awake screaming 
    for hours
S2: \textbf{oh}
S1: so that took its toll
\end{alltt}
\end{framed}
Table 2: Overview of minimal particle types

\subsection{Non-lexical and turn-taking features}

Next, we move beyond lexis and examine non-lexical vocalizations and a range of other aspects of turn-taking such as laughter, pauses, overlaps, and truncation. Our dataset contains non-lexical vocalizations of different functions, such as the minimal particles (also known as interjections) ``hm", ``mhm", ``hmm", ``er", ``erm", ``um", ``aha", ``oh". In fact, this type of utterance ranks among the most frequent in the corpus. These utterances can format a wide range of functions that may be relevant to gender and age category prediction. Unfortunately, our corpus does not annotate functional information of these utterance which makes it difficult to consistently group this type of utterance into functional categories in retrospect (\citealt{liesenfeld2019cantonese}). Inspired by \cite{couper2017interactional}, we therefor decided to only group these particles into five broad form-function mappings based on their typical forms. This way we aim to capture at least some functional variety, even though this unfortunately does not accommodate inter-speaker variation. Table 2 shows the functions we differentiate. 

\begin{table*}[!ht]
\begin{center}
\setcounter{table}{2}
\begin{tabular}{|l|l|r|r|r|r|}
\hline
\multicolumn{2}{|c|}{\multirow{3}{*}{\textbf{Feature}}} & \multicolumn{4}{c|}{\textbf{Category}} \\ \cline{3-6}
 
\multicolumn{2}{|c|}{} & \multicolumn{1}{c|}{\textbf{Female}} & \multicolumn{1}{c|}{\textbf{Male}} & \multicolumn{1}{c|}{\textbf{Old}} & \multicolumn{1}{c|}{\textbf{Young}} \\ \hline
\multirow{5}{*}{minimal particles} & \pbox{6cm}{\shortstack[l]{\bf{Positive responses and continuers} \\ (mm, mhm, mm\_hm, aha, uhu, uhuh, \\ uh\_huh) gender n=86,098; age n=10,506 }} & \cellcolor{blue!35} \bf{highest} &  \cellcolor{blue!10} 67.8\% & \cellcolor{blue!25}97.5\% & 54.6\% \\ \cline{2-6} 
 & \shortstack[l]{\bf{Turn stalling (hmm, hmmm)} \\ gender n=2,722; age n=132} &  \cellcolor{blue!35} \bf{highest} & \cellcolor{blue!25} 64.2\% & 25\% & \cellcolor{blue!10} 33.4\% \\ \cline{2-6} 
 & \shortstack[l]{\bf{Turn management (um, er, erm) } \\ gender n=98,442; age n=16,777 } &  62.8\% & \cellcolor{blue!25} 77.3\% & \cellcolor{blue!35}{\bf highest} &  \cellcolor{blue!10}64\% \\ \cline{2-6} 
 & \shortstack[l]{\bf{ Repair initiators (hm?) } \\ gender n=195; age n=15 } &  \cellcolor{blue!35}\bf{highest} & \cellcolor{blue!10}61.9\% & 19\% &  \cellcolor{blue!25}71.4\% \\ \cline{2-6}
 & \shortstack[l]{\bf{Change-of-state token (oh) } \\ gender n=96,566; age n=15,852 } &  \cellcolor{blue!25} 80.6\%  &  62.4\% & \cellcolor{blue!35}\bf{highest}  & \cellcolor{blue!10} 71.2\% \\
 \hline
\multirow{5}{*}{\shortstack{turn-taking \\ properties \\ and other \\ vocalizations }} & \shortstack[l]{\bf{laughter} \\ gender n=92,417; age n=15,603} & \cellcolor{cyan!25} 72.5\% & \cellcolor{cyan!10} 59.1\%  & 58.6\% & \cellcolor{cyan!35} \bf{highest} \\ \cline{2-6} 
 & \shortstack[l]{\bf{pause (short, 1-5 sec)} \\ gender n=236,885; age n=29,703} &  \cellcolor{cyan!35}\bf{highest} & \cellcolor{cyan!10}91.2\% &  \cellcolor{cyan!25}94.4\% & 76.6\% \\ \cline{2-6} 
 & \shortstack[l]{\bf{truncated words} \\ gender n=68,122; age n=11,065 }&  80\% & \cellcolor{cyan!10}89\% & \cellcolor{cyan!35}{\bf highest} &  \cellcolor{cyan!25}91.\% \\ \cline{2-6} 
 & \shortstack[l]{\bf{overlaps (by total turns ratio)} \\ gender n=250,628; age n=46,285} & \cellcolor{cyan!10} 87\% & 80\% &  \cellcolor{cyan!25}90\% &  \cellcolor{cyan!35}{\bf highest} \\ \hline
\end{tabular}
\end{center}
\caption{\label{font-table} Overview of non-lexical features in the dataset: minimal particles, turn-taking properties and other vocalizations (in relative frequencies,  first rank is displayed as ``highest" and rank 2-4 in percentage of first rank, blue and teal intensity indicates rank, n = counts of each feature by subcorpus)}
\label{table:3}
\end{table*}

In addition, laughter, truncation , pauses and overlap is also annotated in our corpus as single labels that indicate the occurrence of laughter-related sounds, abandoned words, as well as the occurrence of overlap between two turns by speakers. Table 3 provides an overview of these non-lexical vocalizations and turn-taking properties.

The cells in dark blue show the highest occurrence of a feature per category as relative frequencies. For example, laughter is most prominent among young speakers, while turn management tokens ("er", "erm", "um") are typical for old speakers. The lighter blue cells compare the prominence of the same item across categories, displayed as the percentage of the highest ranking category.

First we look at minimal particles that typically format positive responses (as for acknowledgments) and continuers. This includes nasal utterances such as ``mm", ``mhm", ``mm\_hm", as well as vocalic utterances such as ``aha", ``uhu", ``uhuh", ``uh\_huh". Turns by female and old speakers feature these utterances more often as those of other speakers. Second, we examine utterances typically related to turn stalling or management. We distinguish two types of forms that typically format this, nasal ``hmm" and ``hmmm" sounds as well as vocalic sounds ``um", ``er", and ``erm". Turn stalling tokens appear most frequently in turns by female speakers while turn management tokens appear predominantly in turns by old speakers. Next, we examine nasal utterances featuring a rising pitch which are annotated as ``hm?". This type of utterance can format doubt, disbelieve, or serve as a repair initiator. It appears predominantly in turns by female speakers (notably raw counts for this token are very low). Lastly, the utterance ``oh", that (as a response and with rising pitch) commonly formats a change-of-state token that expresses an insight or understanding, features most prominently in talk by old speakers.

\section{Prediction}

Can we predict the speaker's gender and age category based on lexical, non-lexical and turn-taking features alone? And how does including non-lexical vocalizations impact the binary classification task?

\subsection{Controlling the data} 

One challenge of working with transcriptions of unscripted conversations is that various subcorpora that one would like to compare are rarely of the same size. For binary classification it is therefore essential to select equal numbers of speakers of each category. We also checked the amount of utterances of each speaker and removed those which only produced a minimal amount of talk. 

Furthermore, we considered controlling for gender pairs, making sure our subcorpora feature male-male, female-male, and female-female talk in equal measure, but ultimately we decided that, in this case, the resulting dataset would not be big enough for the task. Similarly, we decided against using a leave-one-label-out split to control for the language of a particular conversation.

\subsection{Results}
First, we predicted the label of a single speaker based on all their utterances. We obtained 305 speakers each for classifying gender and around 50 each for age. Especially the size of the age corpus is therefore almost unsuitable for a classification task, so we caution the reader to treat the resulting classification accuracy with a grain of salt. Using the features discussed in Section 3, we began with only considering lexical features, and then considering both lexical and non-lexical. We then train/tested (50/50) a Logistic Regression classifier to predict gender and age of each speaker with 10-fold cross-validation. We obtained a classification accuracy of 71\% for gender labels and 90\% for age labels using only lexical features, after added non-lexical features the accuracy increased to 81\% and 92\% respectively.  

Second, we also tried to predict the label of a speaker per individual turn. Similarly, we split the dataset into equal amounts of turns per category (around a million turns for gender, and around 200,000 for age), and trained a classifier using a 90/10 train/test ratio to predict the gender and age label of a single turn. Here, we obtain 62\% accuracy for gender and 67\% for age using lexical features, and 63\% and 67\% after adding non-lexical features. 

In both settings non-lexical features increased the accuracy of the classifier which indicates that this information is a useful discriminator of the labels. 

Third, we also fine-tune a pretrained language model (BERT-large, uncased) and achieve an accuracy of 100\% for predicting both age and gender per speaker, and 89\% and 91\% for per-turn prediction \citep{devlin2019bert}. Notably, BERT's vocabulary does not contain non-lexical utterances which results in the loss of this information. This is unfortunate for per-turn prediction where such items can make up a significant part of the content. Would including non-lexical items increase accuracy here too?

%as well as the area that this study aims to contribute: user profiles and speaker trait detection in voice bots and spoken dialog systems.
\begin{table*}[!ht]
\centering
\begin{tabular}{|l|c|c|}
\hline
 {\textbf{Features}} & \textbf{Gender} & \textbf{Age} \\ 
  & \textbf{Accuracy}$\pm$    \textbf{Std. Error} & \textbf{Accuracy}$\pm$    \textbf{Std. Error} \\ \hline
 \textbf{per conversation} & & \\ \hline
baseline & 50$\pm$ 0 \% & 50$\pm$ 0 \%\\ \hline
lexical & 70.95$\pm$ 5.38 \% & 89.50$\pm$ 3.53 \%\\ \hline
lexical + non-lexical & 80.71$\pm$ 5.62 \% & 92.00$\pm$ 3.27 \%\\ \hline
BERT (lexical only) & 100.00$\pm$ 0.00 \% & 100.00$\pm$ 0.00 \%\\ \hline
\textbf{per turn} & &\\ \hline
baseline & 50$\pm$ 0 \% & 50$\pm$ 0 \%\\ \hline
lexical & 62.39$\pm$ 0.18 \% & 67.33$\pm$ 0.34 \%\\ \hline
lexical + non-lexical & 63.57$\pm$ 0.13 \% & 67.74 $\pm$ 0.46 \%\\ \hline
BERT (lexical only) & 89.36$\pm$ 0.22 \% & 90.53$\pm$ 0.27 \%\\ \hline
\end{tabular}
 \caption{Results of category prediction as a binary classification task of ``male" - ``female", ``young" - ``old" labels, per single-speaker and per turn}
\end{table*}

\section{Conclusion}
\label{ssec:first}

We examined gender and age salience and (stereo)typicality in British English conversation. The results of this pilot study show that a range of lexical, phrasal, non-lexical, and turn-taking-related features exhibit a tendency to appear more prominently across binary gender and old categories. We were especially interested in the use of non-lexical vocalizations, particles, exclamations and other turn-taking dynamics. Here, we found that female speakers produce significantly more and slightly longer turns. Talk by female speakers also tends to feature the minimal particles ``huh", ``hm" and ``mm" more prominently. In contrast, male speakers' talk tends to be characterized by the minimal particles ``eh", ``uh" and ``em". Overall, male speakers tend to produce shorter turns with fewer words and a higher type-token ratio.

Looking at generational differences, we found that young speakers laugh more and their turns overlap more and tends to feature more words typically related to swearing such as ``shit", ``fuck" or ``fucking". Talk by old speakers tends to feature more truncated words and turn management tokens.  
% Notably, several phrases that appear more prominently in female talk also appear in young speakers' talk, while phrases that appear in old speakers' talk also feature more prominently in male speakers' talk. This suggests links between gendered and age-salient talk, such as that gendered performances in talk may contain elements that also feature more prominent in young speakers' turns.

Based on such observations of characteristics across categories, we set up a classification task to predict gender and age labels of both single speakers and individual turns. We found that predicting speaker labels per conversation yields significantly higher classification accuracy in comparison to the prediction of labels for individual turns. This is likely due to the high number of very short turns that don't feature utterances with label bias. The classification results of around 80\% for predicting speaker labels per conversation show that a simple logistic regression classifier does a reasonably good job even when confronted with ``unstructured" and ``messy" transcribed speech. Notably, we show that non-lexical utterances and minimal particles, which are often filtered out in dialog and speech corpus datasets, contribute to more accurate prediction.

\section{Limitations and further studies}

% \indent We would like to highlight the potential conceptual pitfalls of thinking of speaker's gender and age category prediction as a binary classification task. The use of labels for participants can lead to the dissemination of biased conceptions of gender and age salient performances in conversation. \\ 
% We would like to stress the need to be very cautious when making inferences based on data labelled for social categories such as used in this study. No study on related topics can be a study on computational modelling eo ipso. This underscores the need for more inclusive language resources in the area. \\
\indent Despite the significant conceptual pitfalls that come with labelling participants for gender and age categories, we hope that our preliminary results yielded some interesting insights on gender and age (stereo)typicality in contemporary British English talk and will draw more attention to much-needed computational work based on authentic, real-world recordings instead of sterile, polished datasets.    

In the real world, gender and age performances in talk-in-interaction are not classification tasks. Challenges for the big data approach to user modeling are plenty. For instance, a more comprehensive model needs to take into account that speakers perform gender and age differently across various conversational settings. When more datasets become available, a natural extension to the existing prediction study would be explorations of differences across various conversational compositions. Would we observe similar patterns in conversations with speakers of the same or different gender and age? \\
\indent Another important extension to the current type of study are more detailed explorations of turn-taking dynamics that look into more fine-grained aspects of different types of actions in conversation. User modeling as a text classification problem yields good results for broad categories. But humans are often able to make informed guesses on very specific speaker traits based on style or format of just one turn. Modelling this requires a more sophisticated model of turn types, conversational moves, and the fine-grained systematics of talk-in-interaction. However, quantitative methods are often not well suited to capture more subtle differences of how speakers format various action types, which leads to challenges of how to model the sequential unfolding of action in more detail \citep{liesenfeld2019action}. \\
\indent A critical challenge for the data-driven prediction of gender and age salience in talk is therefore how to take variation in formats of specific actions and activities into account, especially those that have been described as gendered or age-salient such as hedging or ``troubles talk" (\citealt{lakoff1973language}; \citealt{jefferson1988sequential}). Focusing on specific actions would enable a more fine-grained analysis of how speakers negotiate their concepts of gender and age in interaction as part of specific sequences in conversation and how navigating these concepts in interaction relates to (stereo)typical gender and age salience.

%\section*{Acknowledgments}

%Do not number the acknowledgment section. Do not include this section when submitting your paper for review.

\bibliography{paclic34.bib}

\begin{thebibliography}{}

\bibitem[Abouelenien et~al., 2017]{abouelenien2017multimodal}
Abouelenien, M., P{\'e}rez-Rosas, V., Mihalcea, R., and Burzo, M. (2017).
\newblock Multimodal gender detection.
\newblock In {\em Proceedings of the 19th ACM International Conference on
  Multimodal Interaction}, pages 302--311.

\bibitem[Acton, 2011]{acton2011gender}
Acton, E.~K. (2011).
\newblock On gender differences in the distribution of um and uh.
\newblock {\em University of Pennsylvania Working Papers in Linguistics},
  17(2):2.

\bibitem[Argamon et~al., 2003]{argamon2003gender}
Argamon, S., Koppel, M., Fine, J., and Shimoni, A.~R. (2003).
\newblock Gender, genre, and writing style in formal written texts.
\newblock {\em Text \& Talk}, 23(3):321--346.

\bibitem[Baker, 2014]{baker2014using}
Baker, P. (2014).
\newblock {\em Using corpora to analyze gender}.
\newblock A\&C Black.

\bibitem[Bird et~al., 2009]{bird2009natural}
Bird, S., Klein, E., and Loper, E. (2009).
\newblock {\em {Natural language processing with Python: analyzing text with
  the natural language toolkit}}.
\newblock O'Reilly Media, Inc.

\bibitem[Couper-Kuhlen and Selting, 2017]{couper2017interactional}
Couper-Kuhlen, E. and Selting, M. (2017).
\newblock {\em {Interactional linguistics: Studying language in social
  interaction}}.
\newblock Cambridge University Press.

\bibitem[Devlin et~al., 2019]{devlin2019bert}
Devlin, J., Chang, M.-W., Lee, K., and Toutanova, K. (2019).
\newblock Bert: Pre-training of deep bidirectional transformers for language
  understanding.
\newblock In {\em Proceedings of the 2019 Conference of the North American
  Chapter of the Association for Computational Linguistics: Human Language
  Technologies, Vol 1}, pages 4171--4186.

\bibitem[Herring and Paolillo, 2006]{herring2006gender}
Herring, S.~C. and Paolillo, J.~C. (2006).
\newblock Gender and genre variation in weblogs.
\newblock {\em Journal of Sociolinguistics}, 10(4):439--459.

\bibitem[Honnibal and Montani, 2017]{honnibal2017spacy}
Honnibal, M. and Montani, I. (2017).
\newblock {Spacy 2: Natural language understanding with bloom embeddings,
  convolutional neural networks and incremental parsing}.
\newblock {\em To appear}, 7(1).

\bibitem[James and Drakich, 1993]{james1993understanding}
James, D. and Drakich, J. (1993).
\newblock Understanding gender differences in amount of talk: A critical review
  of research.
\newblock In Tannen, D., editor, {\em Oxford studies in sociolinguistics.
  Gender and conversational interaction}, page 281–312. Oxford University
  Press.

\bibitem[Jefferson, 1988]{jefferson1988sequential}
Jefferson, G. (1988).
\newblock On the sequential organization of troubles-talk in ordinary
  conversation.
\newblock {\em Social problems}, 35(4):418--441.

\bibitem[Joshi et~al., 2017]{joshi2017personalization}
Joshi, C.~K., Mi, F., and Faltings, B. (2017).
\newblock Personalization in goal-oriented dialog.
\newblock In {\em NIPS 2017 Conversational AI Workshop, 4-9 Dec 2017}.

\bibitem[Kessler, 2017]{kessler2017scattertext}
Kessler, J.~S. (2017).
\newblock {Scattertext: a Browser-Based Tool for Visualizing how Corpora
  Differ}.
\newblock In {\em Proceedings of ACL-2017 System Demonstrations, 30 July - 4
  August 2017}, Vancouver, Canada. Association for Computational Linguistics.

\bibitem[Lakoff, 1973]{lakoff1973language}
Lakoff, R. (1973).
\newblock Language and woman's place.
\newblock {\em Language in society}, 2(1):45--79.

\bibitem[Liesenfeld, 2019a]{liesenfeld2019action}
Liesenfeld, A. (2019a).
\newblock {\em {Action formation with janwai in Cantonese Chinese
  conversation}}.
\newblock PhD thesis, Nanyang Technological University.

\bibitem[Liesenfeld, 2019b]{liesenfeld2019cantonese}
Liesenfeld, A. (2019b).
\newblock {Cantonese turn-initial minimal particles: annotation of
  discourse-interactional functions in dialog corpora}.
\newblock {\em Proceedings of the 33rd Pacific Asia Conference on Language,
  Information and Computation (PACLIC 33)}.

\bibitem[Liesenfeld and Huang, 2020]{Liesenfeld}
Liesenfeld, A. and Huang, C.~R. (2020).
\newblock {NameSpec Asks: What's Your Name in Chinese? A Voice Bot to Specify
  Chinese Personal Names through Dialog}.
\newblock In {\em Proceedings of the 2nd Conference on Conversational User
  Interfaces}, CUI '20, New York, NY, USA. Association for Computing Machinery.

\bibitem[Love et~al., 2017]{love2017spoken}
Love, R., Dembry, C., Hardie, A., Brezina, V., and McEnery, T. (2017).
\newblock {The Spoken BNC2014: Designing and building a spoken corpus of
  everyday conversations}.
\newblock {\em International Journal of Corpus Linguistics}, 22(3):319--344.

\bibitem[McEnery and Xiao, 2004]{mcenery2004swearing}
McEnery, A. and Xiao, Z. (2004).
\newblock {Swearing in modern British English: the case of fuck in the BNC}.
\newblock {\em Language and Literature}, 13(3):235--268.

\bibitem[Oger, 2019]{oger2019study}
Oger, K. (2019).
\newblock {A Study of Non-Finite Forms of Anaphoric do in the Spoken BNC}.
\newblock {\em Anglophonia. French Journal of English Linguistics}, 28.

\bibitem[Prabhakaran and Rambow, 2017]{prabhakaran2017dialog}
Prabhakaran, V. and Rambow, O. (2017).
\newblock Dialog structure through the lens of gender, gender environment, and
  power.
\newblock {\em Dialogue \& Discourse}, 8(2):21--55.

\bibitem[Schofield and Mehr, 2016]{schofield2016gender}
Schofield, A. and Mehr, L. (2016).
\newblock Gender-distinguishing features in film dialogue.
\newblock In {\em Proceedings of the Fifth Workshop on Computational
  Linguistics for Literature, 16 June, 2016}, pages 32--39.

\bibitem[Skantze, 2017]{skantze2017predicting}
Skantze, G. (2017).
\newblock {Predicting and regulating participation equality in human-robot
  conversations: Effects of age and gender}.
\newblock In {\em 2017 12th ACM/IEEE International Conference on Human-Robot
  Interaction (HRI), 9-11 March, 2017}, pages 196--204. IEEE.

\bibitem[Tannen, 1990]{tannen1990you}
Tannen, D. (1990).
\newblock {\em {You just don't understand: Women and men in conversation}}.
\newblock Morrow New York.

\bibitem[Tannen, 1993]{tannen1993gender}
Tannen, D. (1993).
\newblock {\em Gender and conversational interaction}.
\newblock Oxford University Press.

\bibitem[Wolters et~al., 2009]{wolters2009age}
Wolters, M., Vipperla, R., and Renals, S. (2009).
\newblock {Age recognition for spoken dialogue systems: Do we need it?}
\newblock In {\em Tenth Annual Conference of the International Speech
  Communication Association, 6-10 September 2009}.

\bibitem[Xiao and Tao, 2007]{xiao2007corpus}
Xiao, R. and Tao, H. (2007).
\newblock {A corpus-based sociolinguistic study of amplifiers in British
  English}.
\newblock {\em Sociolinguistic studies}, 1(2):241--273.

\end{thebibliography}

\end{document}